\documentclass[11pt,a4paper]{article}

\usepackage[utf8]{inputenc}
\usepackage[T1]{fontenc}
\usepackage{lmodern}
\usepackage{microtype}
\usepackage[
    top=2.4cm,
    bottom=2.4cm,
    left=2.5cm,
    right=2.5cm
]{geometry}

\usepackage{amsmath,amssymb,mathtools}
\usepackage{booktabs}
\usepackage{array}
\usepackage{tabularx}
\usepackage{enumitem}
\usepackage{graphicx}
\usepackage{float}
\usepackage{fancyhdr}
\usepackage{lastpage}
\usepackage{titlesec}
\usepackage{xcolor}
\usepackage{cite}
\usepackage{hyperref}

\newcommand{\reporttitle}{Avatar Intelligence}
\newcommand{\coverreporttitle}{Avatar Intelligence: Amplifying Human Agency Through Persistent Personal Agents}
\newcommand{\reportsubtitle}{Amplifying Human Agency Through\\Persistent Personal Agents}

\newcommand{\reportversion}{Version 1.0}
\newcommand{\reportdate}{\today}
\newcommand{\reportauthor}{Zheng Wu, Xinbei Ma, Zhaoyuan Huang, Yao Yao, Haowen Hu, Yijie Lu, Taolei Wang, Yifei Yang, Wuyue Xiong, Pengzhou Cheng, Zhuosheng Zhang}
\newcommand{\reportauthorblock}{%
    Zheng Wu\textsuperscript{1,*}, Xinbei Ma\textsuperscript{2,*},
   Yao Yao\textsuperscript{1}, Zhaoyuan Huang\textsuperscript{1}\\
    Haowen Hu\textsuperscript{1}, Yijie Lu\textsuperscript{1},
    Taolei Wang\textsuperscript{1}, Yifei Yang\textsuperscript{1}\\
    Wuyue Xiong\textsuperscript{1}, Pengzhou Cheng\textsuperscript{3},
    Zhuosheng Zhang\textsuperscript{1,\ensuremath{\dagger}}%
}
\newcommand{\reportaffiliation}{%
    \textsuperscript{1}Shanghai Jiao Tong University\\
    \textsuperscript{2}National University of Singapore\\
    \textsuperscript{3}Shanghai University%
}
\newcommand{\reportauthornotes}{%
    \textsuperscript{*}Equal contribution.\quad
    \textsuperscript{\ensuremath{\dagger}}Corresponding author.%
}
\newcommand{\coverabstract}{
Personal AI agents are increasingly capable of handling long-horizon tasks across applications, yet users often need to invest substantial effort in specifying goals, resolving ambiguity, and verifying outcomes.
We argue that progress should be measured by how effectively agents amplify human agency: enabling people to pursue their goals with less routine supervision while retaining control over consequential decisions. 
We advocate \textbf{Avatar Intelligence}, a paradigm of persistent personal agents that represent users' goals, values, and constraints through sustained decision delegation across contexts and time.
We describe a maturity roadmap from human-directed execution to autonomous operation within delegated domains, and introduce the Agentic Oversight Index (AOI) as a framework for evaluating delivered value against agent-side operational cost and human oversight burden.
We identify three core research challenges: identity-continuous personalization that carries user preferences across tasks, user-situated evolution through Rehearsal Engineering, and safety mechanisms that combine technical control with accountability and recourse. 
These directions define a research agenda for reliable decision delegation to personal agents across tasks and over time.
}

\definecolor{ReportNavy}{HTML}{12263A}
\definecolor{ReportBlue}{HTML}{2563EB}
\definecolor{ReportCyan}{HTML}{00A6A6}
\definecolor{ReportInk}{HTML}{1F2937}
\definecolor{ReportGray}{HTML}{6B7280}
\definecolor{ReportLight}{HTML}{F3F6FA}
\definecolor{ReportRule}{HTML}{D9E2EC}

\hypersetup{
    colorlinks=true,
    linkcolor=ReportBlue,
    urlcolor=ReportBlue,
    citecolor=ReportCyan,
    hypertexnames=false,
    pdfauthor={\reportauthor},
    pdftitle={\coverreporttitle}
}

\setlist[itemize]{leftmargin=1.5em,itemsep=0.25em,topsep=0.25em}
\setlist[enumerate]{leftmargin=1.6em,itemsep=0.25em,topsep=0.25em}

\titleformat{\section}
    {\Large\bfseries\color{ReportNavy}}
    {\thesection}{0.75em}{}
\titleformat{\subsection}
    {\large\bfseries\color{ReportInk}}
    {\thesubsection}{0.75em}{}
\titleformat{\subsubsection}
    {\normalsize\bfseries\color{ReportInk}}
    {\thesubsubsection}{0.75em}{}

\renewcommand{\headrulewidth}{0.4pt}
\renewcommand{\headrule}{\hbox to\headwidth{\color{ReportRule}\leaders\hrule height \headrulewidth\hfill}}

\newcolumntype{L}[1]{>{\raggedright\arraybackslash}p{#1}}
\newcolumntype{Y}{>{\raggedright\arraybackslash}X}

\begin{document}

\begin{titlepage}
    \thispagestyle{empty}
        \begin{minipage}{\textwidth}
        \raisebox{-0.9\height}{\includegraphics[width=5cm]{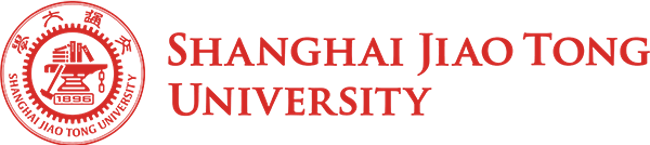}}%
        \hspace{0.15cm}
        \raisebox{-0.9\height}{\includegraphics[
            width=3cm,
            trim={0 196bp 0 197bp},
            clip
        ]{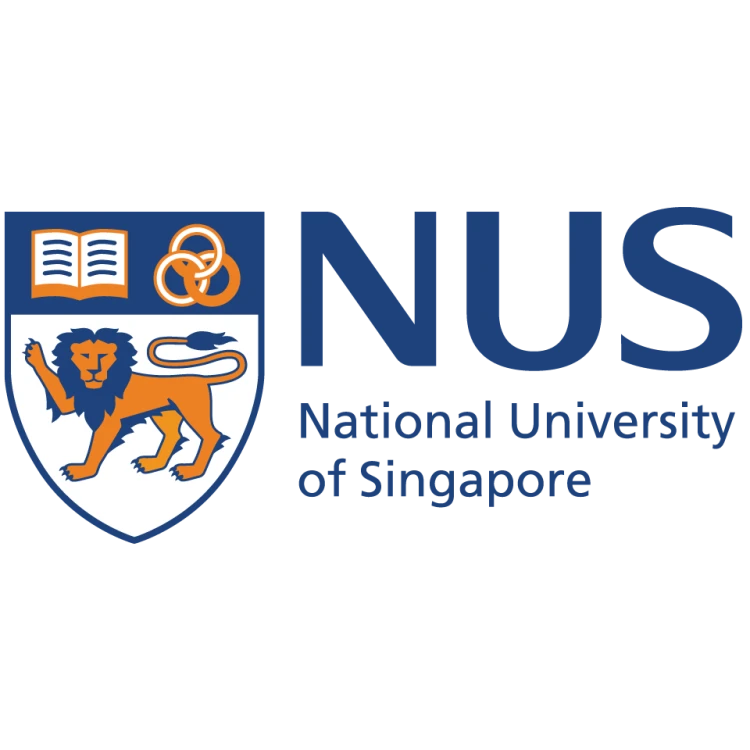}}
        \hspace{0.1cm}
        \raisebox{-0.9\height}{\includegraphics[width=3.8cm]{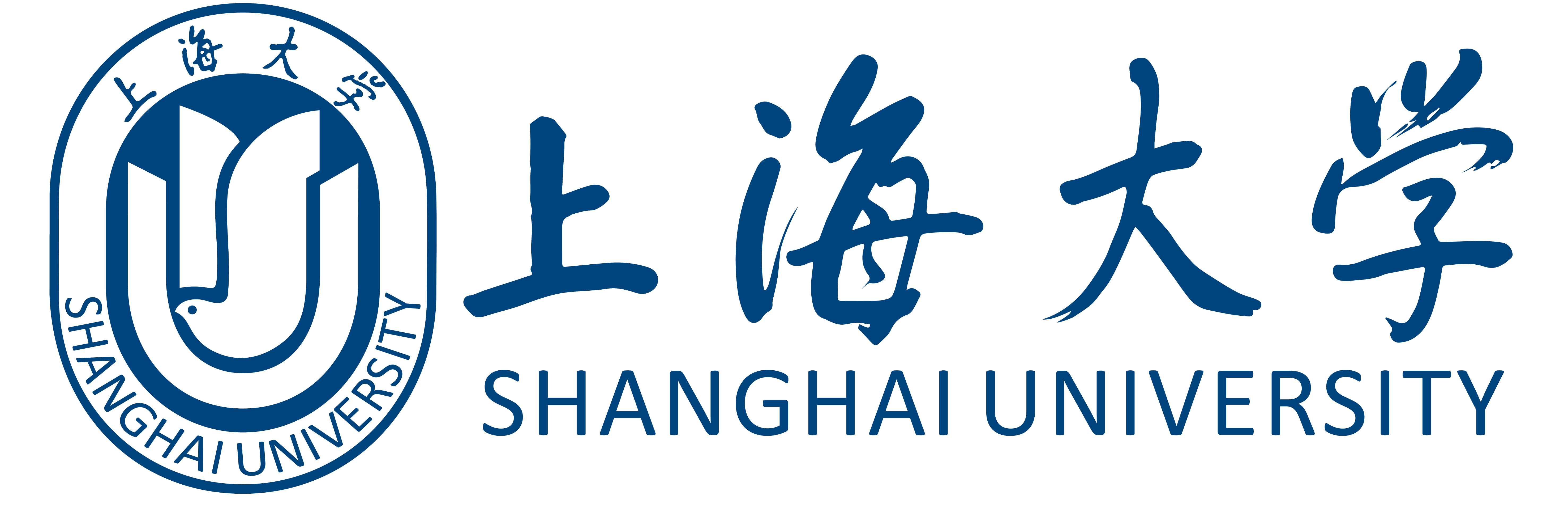}}
    \end{minipage}

    {\color{ReportInk}\rule{\textwidth}{0.5pt}}

    \vspace{0.75cm}
    \begingroup
    \setlength{\fboxsep}{20pt}
    \colorbox{ReportLight}{%
        \begin{minipage}{0.91\textwidth}
            \vspace{0.2cm}
            {\raggedright\color{ReportNavy}
                {\bfseries\fontsize{23}{28}\selectfont\reporttitle:\par}
                \vspace{0.15cm}
                {\bfseries\fontsize{17}{22}\selectfont\reportsubtitle\par}
            }

            \vspace{0.75cm}
            {\small\bfseries \reportauthorblock}\\[0.35em]
            {\small\reportaffiliation}\\[-0.1em]
            {\footnotesize\reportauthornotes}

            \vspace{0.65cm}
            {\small\textbf{Abstract:} \coverabstract\par}

            \vspace{0.45cm}

            \vspace{0.45cm}
            {\small\textbf{Version:} \reportversion \qquad
            \textbf{Date:} \reportdate}

            \vspace{0.55cm}
        \end{minipage}%
    }
    \endgroup

    \vfill
    {\color{ReportInk}\rule{0.40\textwidth}{0.5pt}}
\end{titlepage}

\pagenumbering{roman}

\tableofcontents
\newpage

\pagenumbering{arabic}

\section{Introduction}

\begin{figure}[t]
    \centering
    \includegraphics[
        width=\linewidth,
        trim={0 1cm 0 1cm},
        clip
    ]{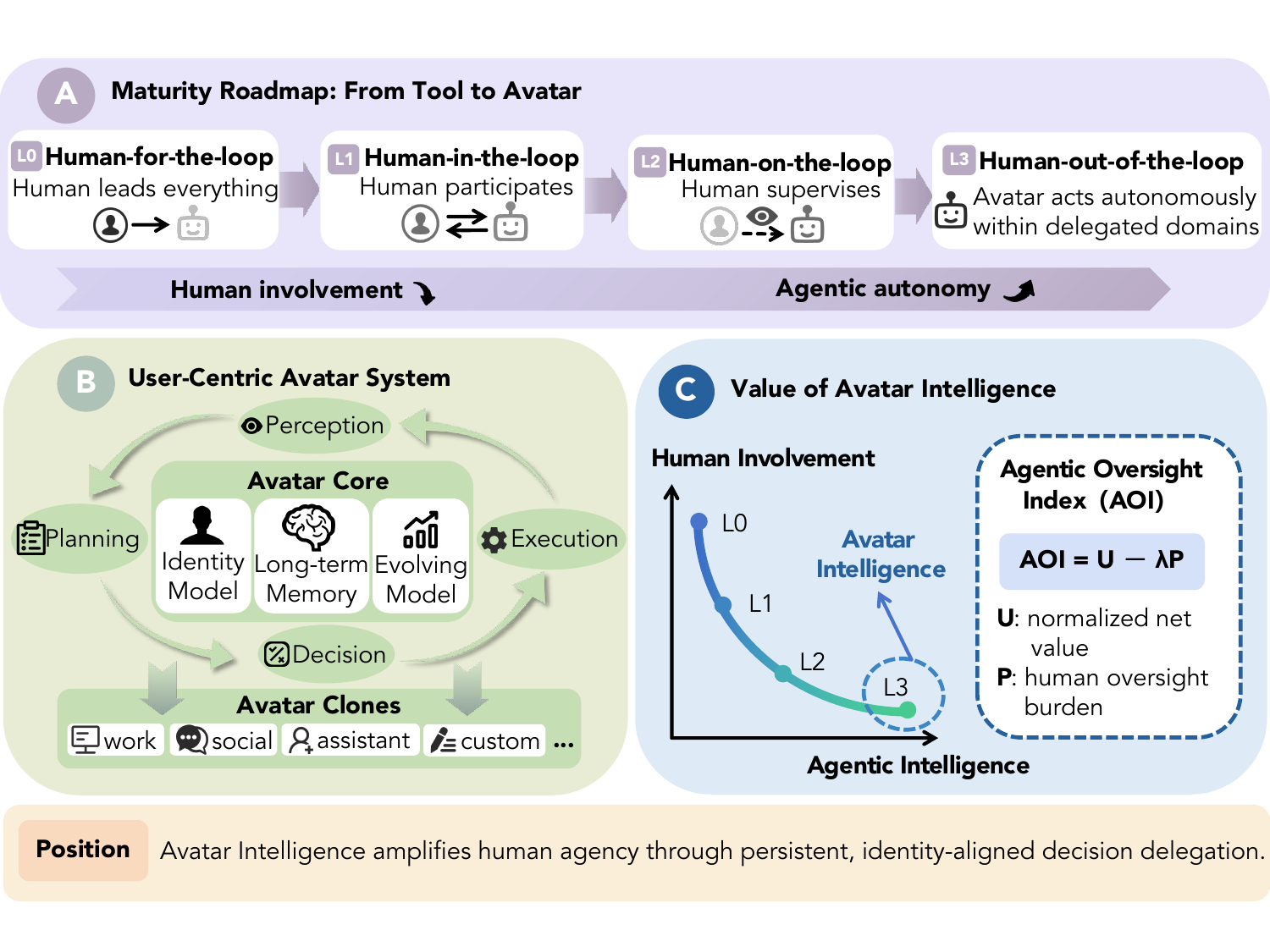}
    \caption{Conceptual overview of Avatar Intelligence. \textbf{(A)} The maturity roadmap traces the transfer of execution and decision responsibilities from users to avatars within delegated domains. \textbf{(B)} A shared avatar core supports a perception--planning--decision--execution loop and task-specialized clones. \textbf{(C)} Evaluation accounts for both delivered value and the human effort required to obtain it: the Agentic Oversight Index, $AOI=U-\lambda P$, combines normalized net value $U$ with an explicit penalty for human oversight burden $P$, weighted by $\lambda$.}
    \label{fig:overview}
\end{figure}

Imagine an AI agent system that acts on your behalf: it makes decisions, executes actions, and communicates with humans or other agents with minimal human intervention. Such a system is more than a tool that waits for explicit commands; it resembles a user-centric operating system---a persistent digital representative that embodies your preferences, values, and goals and operates continuously across contexts and time.

Recent years have witnessed the rapid development of foundation models (e.g., GPT-5.5~\cite{openaigpt5.5}, Gemini-3~\cite{gemeni}, and DeepSeek-V4~\cite{deepseek}), which establish a general-purpose backbone for autonomous agents such as MUSE~\footnote{\url{https://ai.meta.com/muse/}.} and dots~\footnote{\url{https://openai.com/index/introducing-dots/}.}.
Meanwhile, agent frameworks (e.g., those equipped with prompt/context/harness engineering and memory systems) have enabled more robust long-horizon task capabilities~\cite{zhou2026externalization}. Beyond the exchange of information in digital environments, agents can increasingly manipulate end-user devices such as computers and smartphones~\cite{zhang2023you}.

These developments make persistent personal agents increasingly plausible, but leave a central question unresolved: how much work must users still do to make delegation effective? 
When delegating a research task, for example, an agent may retrieve and summarize documents while the user continues to resolve ambiguous objectives, correct interpretations, and judge whether the result is sufficient. 
Execution effort decreases, while the burden of specification, coordination, and verification remains substantial.

Our position is that personal agents should amplify human agency by allowing users to pursue more of their goals without supervising every intermediate decision. 
Human agency includes the ability to set and revise goals, choose which decisions to delegate, and redirect action when circumstances change. 
Progress in personal agents therefore depends on reducing the effort required to obtain reliable, user-aligned outcomes while preserving these forms of control.

We use \textbf{Avatar Intelligence} to describe persistent personal agents that represent a user's goals, values, and constraints across tasks and over time. 
Concretely, an avatar should maintain commitments while the user is unavailable,  adapt to changing circumstances, and know when to seek clarification or additional authorization.
This requires learning from ongoing interactions and evaluating how reliably the agent represents the user over time.


We develop this position through four contributions:
\begin{itemize}
    \item We characterize Avatar Intelligence through persistent user representation and sustained decision delegation, and describe a user-centric system architecture that supports these functions.
     \item We outline a maturity roadmap (L0--L3) for transferring operational and decision responsibilities within defined delegation scopes.
    \item We introduce the Agentic Oversight Index (AOI), an evaluation framework that accounts for delivered value, agent-side operational cost, and human oversight burden.
    \item We identify three core research directions: identity-continuous personalization, user-situated evolution through rehearsal, and safety mechanisms for controlled and accountable delegation.
\end{itemize}

\section{Avatar Intelligence}
The essence of Avatar Intelligence is to align AI agents with human needs and expectations.
At its core, Avatar Intelligence reshapes the AI--human relationship by transforming agents from passive instruments into autonomous digital representatives that embody user identity, preferences, and values.
Rather than merely automating isolated tasks, avatars extend human agency beyond biological constraints by enabling persistent, autonomous operation across contexts and time.

\subsection{Avatar as a User-Centric System}

\begin{figure}[t]
    \centering
    \includegraphics[width=\linewidth,
        trim={0 1cm 0 1cm},
        clip]{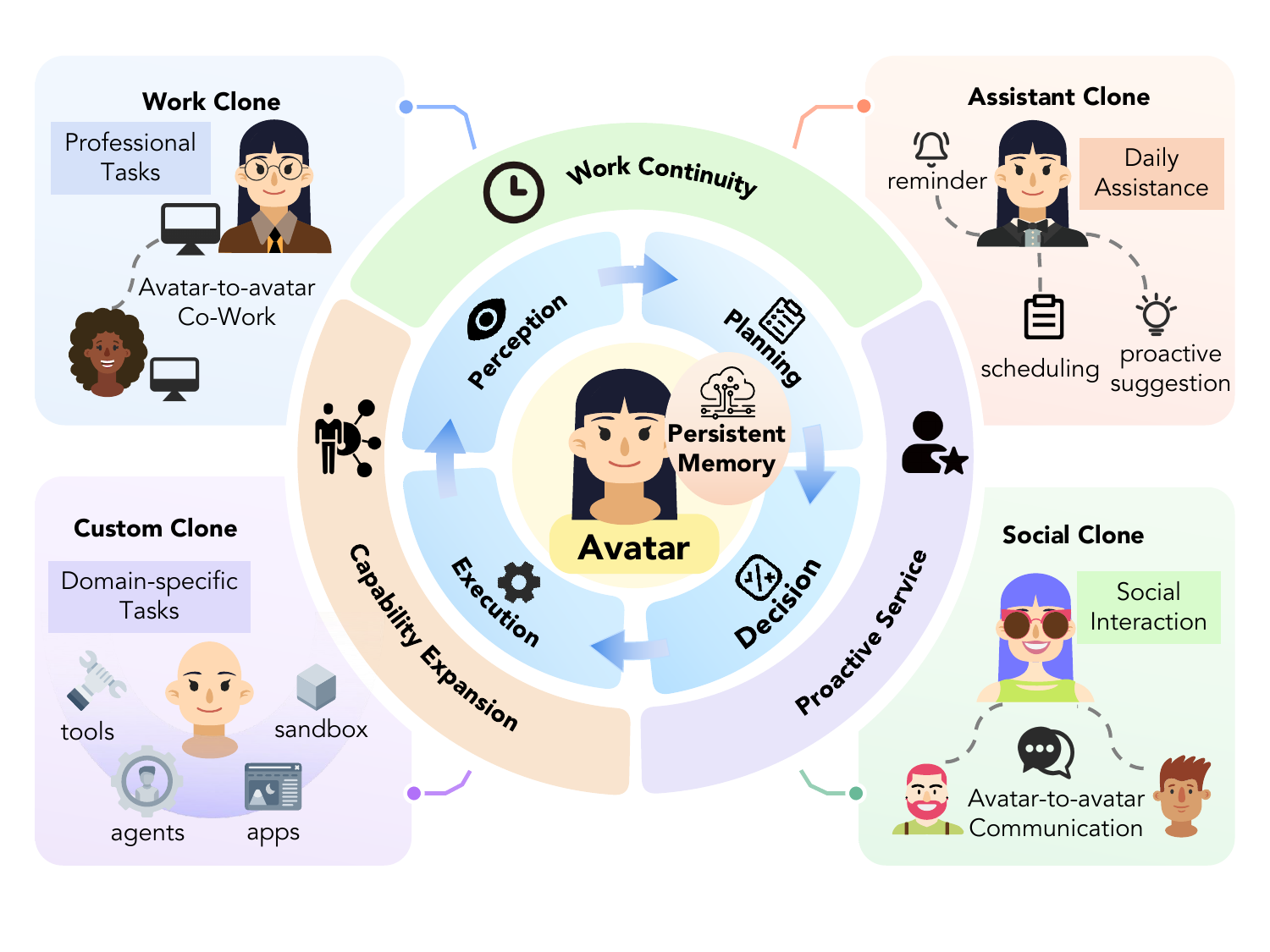}
    \caption{Avatar Intelligence as a user-centric operating system. A persistent user model and memory support the inner loop of perception, planning, decision, and execution. The outer ring identifies three system capabilities: work continuity, proactive service, and capability expansion. Work, assistant, social, and custom clones carry the shared user identity into distinct task contexts, connecting the avatar to tools, applications, and other agents.\label{fig:avatar}}
\end{figure}

We conceptualize Avatar Intelligence as a \textbf{user-centric operating system} that externalizes a user\textquotesingle s cognition into a persistent and executable digital form (Figure~\ref{fig:avatar}). The kernel of such a system is not a thin memory cache, but a faithful and continuously updated model of the user\textquotesingle s cognitive identity.
More specifically, at the OS level, the Avatar continuously observes the user’s behavioral patterns and decision-making logic, implicitly modeling and iteratively refining its understanding of the user’s profile and behavioral patterns. Based on this evolving user model, it can automatically align with the user’s implicit intentions when carrying out tasks on the user’s behalf, thereby enabling more personalized and cognitively faithful task execution.

Existing personalization methods largely optimize behavioral outputs using both training-time methods~\cite{DBLP:conf/nips/ChristianoLBMLA17,yao2020rlper,DBLP:conf/nips/LeePKS24,P-RLHF,jin2025tagging} and decoding-time alignment~\cite{chenpad,DBLP:conf/emnlp/KimLJLTJ25,bu2025personalized}, while current memory systems mainly extend context and accumulate experience~\cite{tan2025prospect,kang2025memory,sun2025preference}.

However, current methods are still insufficient for an Avatar that must preserve identity across contexts and over time. Operationally, the Avatar is organized around four tightly coupled components: \textit{perception}, \textit{planning}, \textit{decision}, and \textit{execution}, grounded by a persistent memory substrate. Together, they form a closed loop that transforms environmental signals into user-aligned action, enabling cognition to operate beyond explicit interaction.

In practical deployment, the Avatar instantiates multiple task-specialized clones as contextual projections of a single underlying identity, including work, assistant, social, and custom clones. These clones inherit the same user model while interfacing with heterogeneous execution substrates such as tools, applications, sandboxed environments, and external agents. This clone layer gives the Avatar three system-level capabilities essential to Avatar Intelligence:

\paragraph{Work Continuity} allows task state to persist across time, applications, and contexts, making long-horizon execution possible without repeated user specification.
\paragraph{Proactive Service} enables the system to act on inferred intent rather than waiting for explicit invocation, shifting the paradigm from reactive response to continuous assistance.
\paragraph{Capability Expansion} allows clones to dynamically incorporate new tools, services, and agents, turning the Avatar into an open-ended system whose functional boundary evolves with the user's needs.

As these clones persist and scale, they will increasingly interface with other users' agents, leading to a \textbf{multi-avatar ecosystem} in which many routine digital interactions are mediated by delegated representatives rather than direct human operation. In such settings, the core unit of interaction shifts from \emph{human--system} to \textbf{avatar--avatar} coordination: scheduling, negotiation, information exchange, and joint task execution can occur through policies that represent distinct users' identities and constraints. This shift reframes the technical problem from single-agent task completion to the design of protocols, incentives, and safety boundaries that enable cooperation under conflict, accountability under delegation, and trust under partial autonomy.

\subsection{Roadmap}
\begin{figure}[t]
    \centering
    \includegraphics[width=\linewidth]{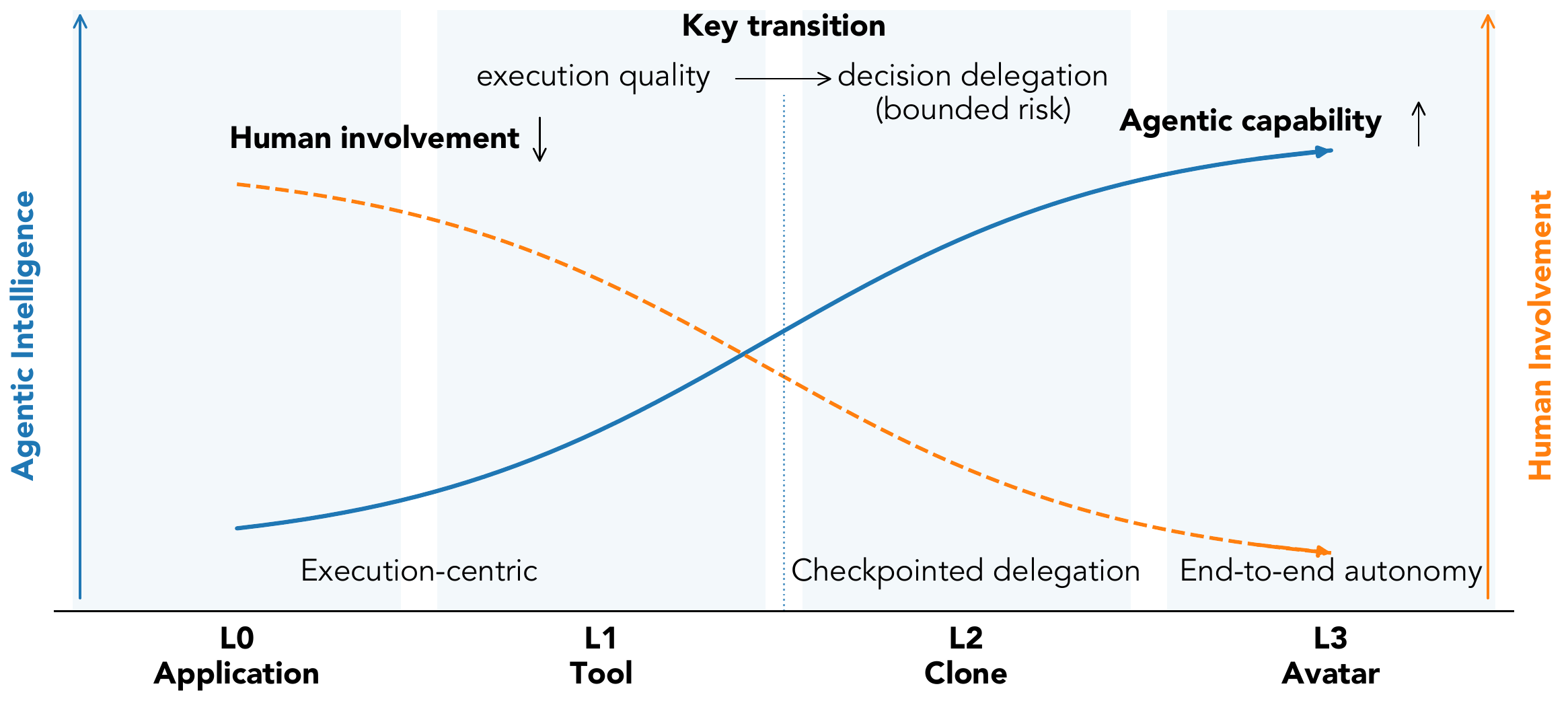}
    \caption{Maturity roadmap toward Avatar Intelligence. The blue solid and orange dashed curves schematically illustrate increasing agentic intelligence and decreasing human involvement across L0--L3. The central transition is from execution-centric assistance to checkpointed decision delegation and ultimately end-to-end autonomy within delegated domains. The same avatar can operate at different levels across domains; users retain authority to revise goals and revoke delegation at every level.}
    \label{fig:Roadmap}
\end{figure}

We view the path from today\textquotesingle s agents to full Avatar Intelligence as a staged decoupling of \emph{human involvement} from \emph{agentic capability} (Figure~\ref{fig:Roadmap}). The stages differ not only in autonomy, but also in the required system guarantees (identity continuity, controllability, and accountability):
\begin{itemize}
    \item \textbf{L0: Human-for-the-loop.} The agent behaves as an augmented application, automating fragmented subprocesses while the user remains embedded in both execution and decisions.
    \item \textbf{L1: Human-in-the-loop.} The agent executes longer workflows, but users still steer key decisions (goal refinement, routing, validation) and absorb most accountability.
    \item \textbf{L2: Human-on-the-loop.} An \emph{avatar core} (persistent identity + memory + decision policy) takes over routine decisions and orchestration; humans intervene at a small number of calibrated checkpoints, with robust runtime guardrails and audit trails.
    \item \textbf{L3: Human-out-of-the-loop.} The avatar assumes end-to-end autonomy over both operations and decisions for delegated domains; humans primarily perform post-hoc auditing, dispute resolution, and policy updates.
\end{itemize}

These levels describe autonomy within a delegated domain, and the same avatar can operate at different levels across domains. Users retain the authority to revise goals and revoke delegation at every level.

This roadmap highlights a key research transition: moving from improving execution quality (L0/L1) to engineering \emph{decision delegation} with bounded risk (L2), and ultimately \emph{institutional-grade assurance} for high-stakes autonomy (L3).

\subsection{Agentic Oversight Index (AOI)}
\label{sec:aoi}
To evaluate progress toward Avatar Intelligence, we argue that capability improvements should be assessed jointly with the \emph{oversight burden} they impose on users. Motivated by the roadmap in Figure~\ref{fig:Roadmap}, we summarize an agent\textquotesingle s maturity by two architectural axes: \textbf{Agentic Intelligence} (how well it executes) and \textbf{Human Involvement} (how much a user must steer, validate, and audit).

\paragraph{Definition.} We introduce the \emph{Agentic Oversight Index (AOI)} as an \emph{oversight-adjusted net-value} score, rather than a ratio: human involvement should penalize a weak agent as well as a strong one. For an avatar $A$ on a task distribution $\tau$, let $U(A,\tau)\in(-1,1)$ denote normalized net value and let $P(A,\tau)\in[0,\infty)$ denote normalized human oversight burden relative to a human-only baseline. We define
\begin{equation}
    U(A,\tau) = \tanh\!\left(
    \frac{V(A,\tau)-C_{\mathrm{sys}}(A,\tau)}{s_{\tau}}
    \right),
\end{equation}
\begin{equation}
    AOI(A,\tau) = U(A,\tau) - \lambda P(A,\tau), \qquad \lambda > 0.
    \label{eq:aoi}
\end{equation}
where $V$ denotes delivered user value (e.g., task success, quality, and user-perceived utility); $C_{\mathrm{sys}}$ denotes agent-side operational cost (e.g., compute and tool usage); and $s_{\tau}>0$ is a task-specific reference scale used to normalize net value. The delivered value $V$ reflects the quality and utility of the outcome, while $C_{\mathrm{sys}}$ accounts for agent-side resource consumption. Crucially, $C_{\mathrm{sys}}$ excludes human oversight, which is represented exclusively by $P$ (e.g., intervention frequency, supervision time, decision complexity, and accountability load). The weight $\lambda$ specifies the value penalty assigned to one unit of normalized oversight burden.

The score is upper-bounded by $1$ but deliberately has no finite lower bound: an agent that creates arbitrarily large human workload should not be made to look merely equally bad. Holding delivered net value fixed, $\partial AOI/\partial P=-\lambda<0$: additional human oversight always lowers the score, including when the avatar has negative net value. When $P\rightarrow0$, AOI converges to $U$ rather than diverging.

\noindent Appendix~\ref{app:aoi} provides one possible instantiation of $U$ and $P$; other operationalizations are compatible with the same principle.

\section{Applications}

\subsection{Current Industry Products}
A new generation of personal agents is beginning to turn persistent assistance into sustained action. Grok Bot keeps named bots and their context across tasks, allowing them to work through applications and return to the user when approval is needed~\cite{grokbot2026}. Muse pursues ongoing goals through connected applications and a persistent browser, while seeking permission for actions such as sending messages or making purchases~\cite{metamuse2026}. Cue envisions agents with their own email, phone, computer, and wallet, subject to boundaries set by the user~\cite{cue2026}. OpenAI's dots likewise continue work between conversations, draw on connected applications and memory, and bring decisions requiring human judgment back to the user~\cite{openaidots2026}.

These systems move beyond one-off responses by carrying work forward over time. Yet the user remains central to setting goals, defining authority, and deciding when consequential actions may proceed. In the roadmap of Avatar Intelligence, they mark a transition from \emph{human-in-the-loop} execution toward \emph{human-on-the-loop} oversight within delegated tasks.
\subsection{Application Scenarios}

\begin{figure}[!htbp]
    \centering
    \includegraphics[width=\linewidth]{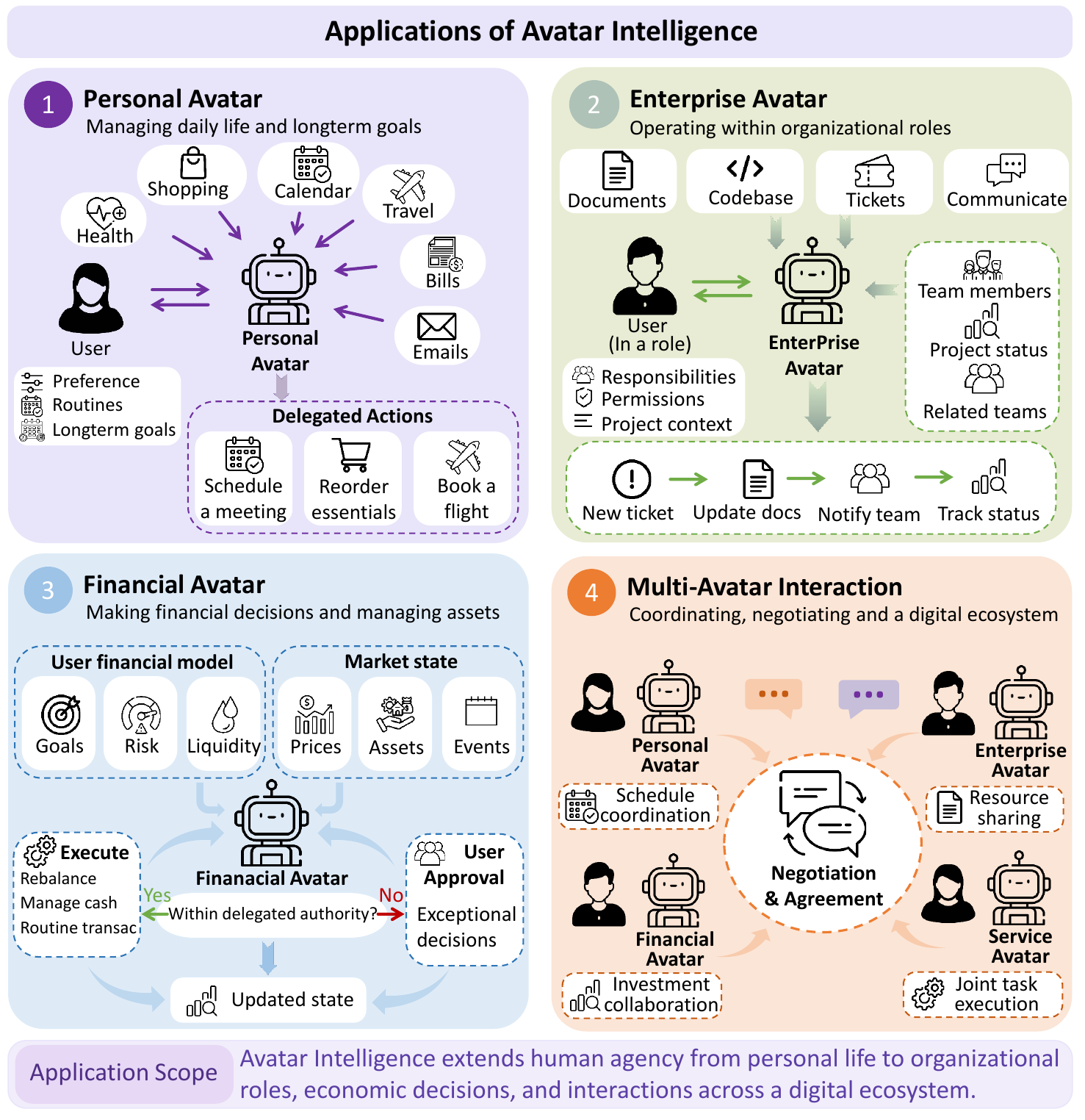}
    \caption{Four application settings for Avatar Intelligence. \textbf{(1)} personal life, \textbf{(2)} organizational work, \textbf{(3)} financial decisions, and \textbf{(4)} coordination among multiple avatars.}
    \label{fig:Applications}
\end{figure}



\paragraph{Personal Avatars for Life Administration and Continuity.}
A personal avatar must carry commitments forward even when they are no longer visible in the immediate request. When a work trip conflicts with a parent--teacher meeting the user promised to attend weeks earlier, the avatar might first look for a later departure rather than simply clearing the calendar. If neither obligation can move, it should bring the conflict back to the user. The value of continuity lies in protecting obligations that a task-by-task assistant would easily overlook.


\paragraph{Enterprise Avatars for Organizational Roles and Operations.}
Suppose engineering flags a release-blocking defect, while the launch calendar and sales brief still promise a release today. Acting for the project manager, an avatar could decide to postpone the internal launch, update the shared schedule, and notify the teams relying on the old date. It cannot waive the engineering blocker or revise a customer commitment on the sales team's behalf; those decisions must return to the people who hold that authority. Such coordination depends on a shared view of the project that changes with new evidence, and on the distinct authority of the roles involved~\cite{vanderhorn2021digital,wu2025colorecosystem}.




\paragraph{Financial Avatars for Economic Decision Making.}
The same market move can call for different actions depending on whose money is at stake. A planned purchase that remains acceptable for one user may push another user's portfolio beyond a chosen exposure limit. In the latter case, the avatar should pause the order and seek a decision, even if the original instruction was to buy. Financial delegation is credible only when the user's risk preferences shape the action itself and changes to those preferences remain under the user's control.


\paragraph{Multi Avatar Interaction and Coordination.}
Once avatars act for different people, a locally sensible decision may be unacceptable to the other party. In arranging a meeting, one avatar may seek the earliest slot while another protects its user's preparation time. They can exchange acceptable windows and settle on a time that respects both constraints, or return to their users if no such time exists. The question is how agents can reach agreements without exposing more of either user's preferences than the negotiation requires.
\section{Open Challenges}
Avatar Intelligence sits at the interface between the user side and the rapidly advancing frontier of agent technology.
On the one hand, an avatar must learn a user’s cognitive and decision patterns—capturing not only preferences, but also reasoning habits, methodologies, and value commitments. As a result, it can reliably understand the user and act as a faithful proxy.
On the other hand, it must continuously grow and evolve in real-world environments: connecting the user to appropriate agent capabilities, simulating the likely consequences of alternative actions, and selecting options that are best aligned with the user’s objectives, thereby building calibrated trust.
Finally, as agents become embedded in everyday life, a central question is how to ensure that such autonomy remains safe and controllable, preventing harm to the user and their environment while preserving the benefits of proactive, long-horizon assistance.

\subsection{Personalization: From Behavioral Alignment to Identity-Continuous Avatars}
\label{sec:personalization}

Personalization is central to Avatar Intelligence: an avatar must make decisions \emph{as the user would}, not merely respond in the user\textquotesingle s preferred style. This requires learning from diverse, authorized signals (instructions, interaction traces, and environmental context) to infer stable decision structure, including values, trade-offs, and reasoning habits.

Most existing approaches~\cite{liu2025learnact,wu2025quick}, however, optimize \emph{behavioral outputs} (what action to take in a particular moment) rather than the latent cognitive model that generates behavior across contexts and over time. In practice, such systems often rely on memory to store user-specific facts or preferences and retrieve them at inference time. While effective for conversational personalization, memory-heavy heuristics can be brittle for agentic tasks: coverage is incomplete, preferences are context-dependent, and user behavior may be internally inconsistent. A core open problem is therefore to model an \emph{identity-continuous} user representation that generalizes across tasks while remaining stable yet updateable.

Complementary to learning-based personalization, many deployed systems also use explicit configuration of identity (values and constraints), memory policies, and skill/tool scopes. A typical example is OpenClaw.~\footnote{\url{https://openclaw.ai/}.} The challenge is to unify these layers into a coherent user model that remains consistent across clones, environments, and time.

We argue that \emph{Avatar Personalization} should be formalized as learning an \emph{identity-continuous} user model that enables transfer and generalization across tasks, while remaining stable yet evolvable.

\begin{figure}[t]
    \centering
    \includegraphics[width=\linewidth,
        trim={0 1cm 0 1cm},
        clip]{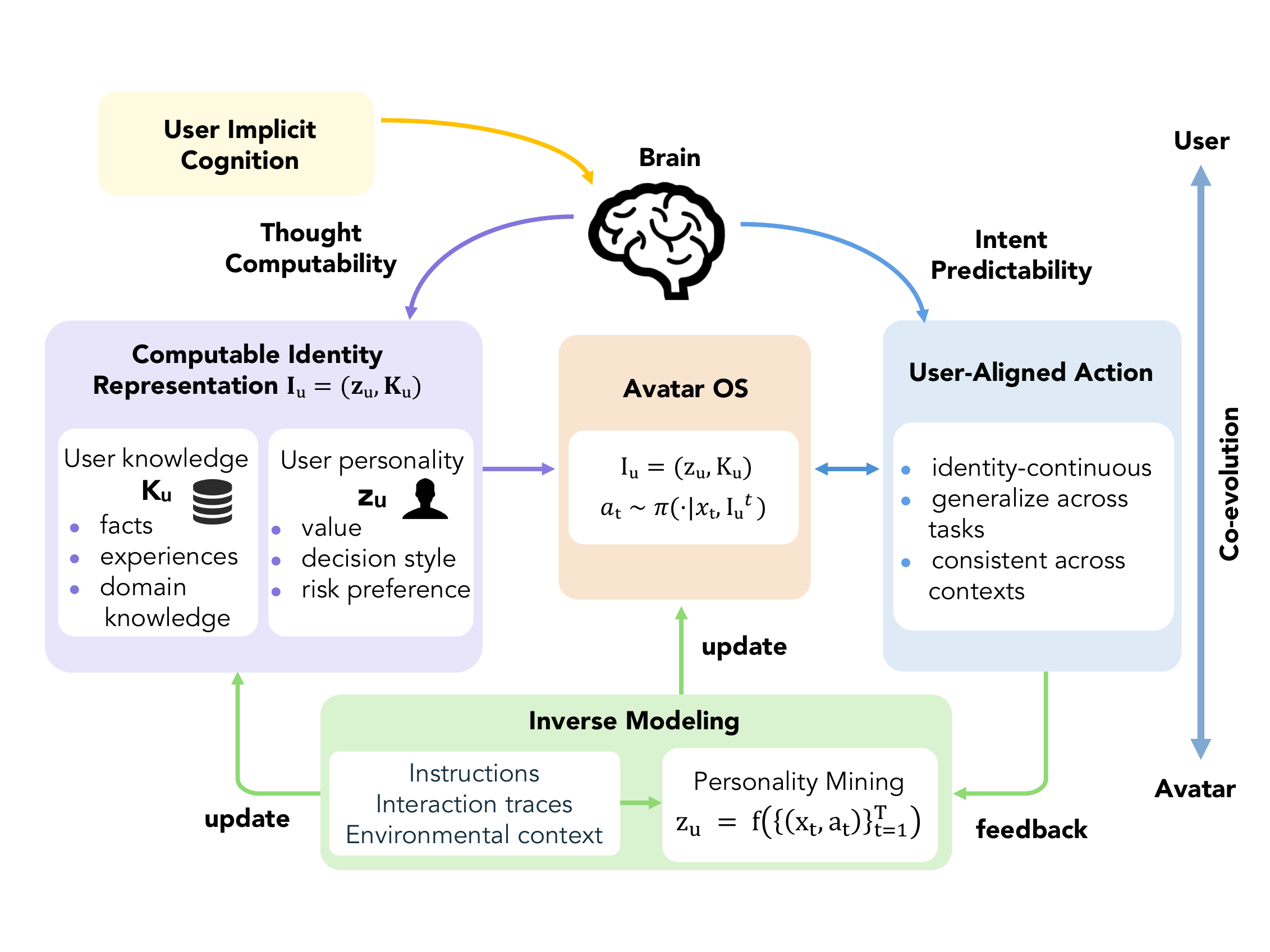}
    \caption{Avatar Personalization through identity modeling and interaction feedback. Thought Computability and Intent Predictability connect implicit user cognition to a computable identity representation, $\mathcal{I}_u=(z_u,\mathcal{K}_u)$, where $z_u$ captures personality structure and $\mathcal{K}_u$ stores user-specific knowledge and experience. Instructions, interaction traces, and environmental context inform inverse modeling of this representation, which guides actions across tasks and contexts. Feedback supports continued refinement of the user model as the user and avatar co-evolve.}
    \label{fig:personalize}
\end{figure}

Figure~\ref{fig:personalize} illustrates how Thought Computability and Intent Predictability support identity-continuous personalization.

\paragraph{From Behavioral Alignment to Computable Identity}
A central requirement for Avatar Intelligence is that the user\textquotesingle s cognition becomes \emph{computable} and \emph{predictable} to the system. Concretely, we characterize Avatar Personalization with two properties:
\begin{itemize}
    \item \textbf{Thought Computability.} The user\textquotesingle s internal decision process admits a compact representation, enabling the avatar to simulate how the user would reason under diverse contexts.
    \item \textbf{Intent Predictability.} Given a context, the avatar can reliably anticipate the user\textquotesingle s intended action without requiring explicit instruction at every step.
\end{itemize}
\noindent Together, these properties shift personalization from surface-level preference matching to learning an \emph{identity-continuous} model that (i) generalizes across tasks, (ii) remains consistent across contexts, and (iii) evolves over time without drifting away from the user\textquotesingle s values and decision structure. We provide a more detailed formalization---including a problem setting, an identity representation, and consistency/stability constraints---in Appendix~\ref{app:formalization}.

\subsection{Evolving with Specialization}
The engineering of autonomous agents has historically focused on improving intent understanding and task completion through external scaffolding. From this perspective, \emph{prompt engineering} injects user intent and constraints into the instruction channel; \emph{context engineering} injects situational state (documents, observations, and memory) to ground planning; and \emph{harness engineering} injects executable rules and control policies (e.g., tool gating, verification, retries, and escalation) to stabilize the action loop. These paradigms primarily shape behavior \emph{at inference time} by conditioning what the agent sees and what it is allowed to do.

\begin{table}[t]
\centering
\small
\setlength{\tabcolsep}{3pt}
\begin{tabularx}{\linewidth}{lYYYY}
\toprule
\textbf{Aspect} & \textbf{Prompt Engineering} & \textbf{Context Engineering} & \textbf{Harness Engineering} & \textbf{Rehearsal Engineering} \\
\midrule
Primary signal & Intent and constraints & Situational state and memory & Runtime control policies and rules & User-situated feedback and outcomes \\
When applied & Per query & Per query / per step & Throughout the action loop & Cross-episode / continual \\
Mechanism & Instruction conditioning & Retrieval and state injection & Orchestration, gating, verification & Scenario staging + evaluation + updates \\
Optimizes for & Specification and style control & Grounding and disambiguation & Reliability and controllability & Specialization and trust calibration \\
Typical risks & Mis-specification & Missing/poisoned context & Brittleness / over-constraint & Drift, privacy, overfitting \\
\bottomrule
\end{tabularx}
\caption{Comparison of four complementary agent engineering paradigms by signal, timing, mechanism, objective, and risk. Prompt engineering specifies intent and constraints; context engineering supplies task-relevant state; harness engineering governs execution; and Rehearsal Engineering uses staged scenarios and user-situated feedback to improve behavior across episodes. Rehearsal provides an adaptation process that draws on, and updates, the other three forms of engineering.}
\label{tab:engineering-paradigms}
\end{table}

A growing challenge, however, is to make agents \emph{evolve} proactively from interaction with the environment. Recent explorations either (i) construct simulators or games where agents can rehearse trajectories, compare alternatives, and improve through iterative critique~\cite{wangvoyager}, or (ii) expose agents to real environments or sandbox~\cite{zhai2025agentevolver,xue2026evocua,xi2406agentgym} where they observe state transitions, discover recurring patterns, and learn from sparse but high-fidelity outcomes. Despite their differences, both lines of work are fundamentally about acquiring \emph{effective feedback}---signals that evaluate decisions, not just executions---and turning such feedback into better future policies.

For Avatar Intelligence, feedback cannot remain purely generic. Beyond learning in standardized benchmarks or public web contexts, an avatar must specialize within the user\textquotesingle s own ecology: their tools, routines, constraints, and risk tolerances. 
We refer to this user-situated specialization loop as \textbf{Rehearsal Engineering}: a system-level methodology for aligning a standardized, language-trained agent to the user\textquotesingle s real environment by (1) collecting user- and environment-specific traces and preferences, (2) staging representative scenarios (via simulation, counterfactual rollouts, or controlled sandboxing), (3) evaluating candidate actions with explicit oversight policies, and (4) updating the avatar\textquotesingle s memory, skill selection, and decision policy under bounded autonomy. 

Situated cognition research in psychology suggests that knowledge is closely tied to the contexts in which it is used~\cite{brown1989situated}. By rehearsing within the user’s own work and life contexts, an agent may learn which contextual cues shape the user’s choices, and how those cues change the user’s trade-offs across situations.
In this view, rehearsal is not merely practice; it is the mechanism by which an avatar internalizes user-specific feedback while remaining safe, auditable, and corrigible.

\subsection{Safety}
Safety is a prerequisite for Avatar Intelligence because avatars are persistent, stateful, and increasingly entrusted with system access and delegated decision authority. We view avatar safety as two complementary layers distinguished by function rather than agent capability. \emph{Technical safety} continuously enforces delegated authority, reduces the likelihood of failures, and limits their consequences. \emph{Agent Insurance} complements these controls by allocating residual financial risk and compensating covered losses under explicit contractual terms. Both layers support delegation across capability levels: greater capability does not remove the need for technical control, and insurance provides defined financial recourse rather than a guarantee of correct or user-aligned execution.

\paragraph{Layer I: Technical safety for continuous control.}
Across capability levels and throughout deployment, safety should be engineered end-to-end around intent fidelity, bounded autonomy, and adversarial robustness:
\begin{itemize}
    \item \textbf{Intent alignment and corrigibility.} Maintain synchronization between the avatar\textquotesingle s policy and the user\textquotesingle s intent through explicit preference/constraint representations, intent-drift detection, and structured escalation (ask/confirm) at high stakes.
    \item \textbf{Runtime governance (harness-level control).} Enforce least privilege, sandboxing, action preconditions, irreversible-action verification, checkpointing/rollback, and deterministic replanning to constrain the action loop even when the model is wrong.
    \item \textbf{Security and integrity hardening.} Continuously monitor tool usage and environment interactions; detect prompt injection and anomalous behaviors; and protect long-term memory, credentials, and dependencies via sanitization and supply-chain scanning.
\end{itemize}

\paragraph{Layer II: Agent Insurance for contractually defined loss compensation.}
Delegated tasks such as fund-moving, code changes, and account operations can expose users to substantial losses when execution fails. Technical safeguards reduce these risks, but residual uncertainty remains in open-world environments. Inspired by recent work on financial risk management for AI agents~\cite{hua2026ars}, we argue that \emph{Agent Insurance}, supported by auditable settlement mechanisms, will become a core infrastructure component for delegation at scale. Service fees can be escrowed and released upon verified delivery; providers can post collateral; and users can optionally purchase underwriting for losses arising from specified failures, such as non-delivery, misexecution, or intent misalignment. Coverage must define verifiable failure conditions, compensation limits, exclusions, and claims procedures. Crucially, compensation requires \emph{auditable evidence}, including execution traces and evaluation artifacts, together with dispute processes. These mechanisms make responsibility for covered losses explicit and provide contractual recourse, while technical controls remain necessary to prevent harm, including harms that financial compensation cannot reverse.

\paragraph{Summary.} Taken together, the open problems in personalization, rehearsal-driven specialization, and safety/assurance constitute the missing infrastructure for advancing from L2 (checkpointed delegation) to L3 (full-pipeline autonomy within delegated domains). Identity-continuous personalization provides a stable decision prior, while Rehearsal Engineering supplies user-situated feedback loops to adapt that prior to real environments. Technical safety continuously constrains delegated actions and limits failure propagation. Agent Insurance provides infrastructure for pricing and allocating residual financial risk and compensating contractually covered losses. Together, continuous technical control and defined financial recourse support trustworthy delegation at scale as avatars assume broader responsibilities with less routine human supervision.


\section{Conclusion}
Avatar Intelligence frames persistent personal agents around a continuing relationship of decision delegation. Its value lies in enabling users to pursue their goals across tasks and over time with less routine supervision and effective control over consequential choices. We outlined a user-centric architecture, a delegation roadmap, and the Agentic Oversight Index to connect this objective with system design and evaluation. The central research agenda is to learn user representations that transfer across contexts, improve decisions through user-situated rehearsal, and support delegation with technical control, accountability, and recourse. Progress should be demonstrated through the decisions users can reliably delegate and the effort required to obtain outcomes they endorse.

\newpage
\bibliographystyle{IEEEtran}
\bibliography{references}

\appendix

\section{Formalization of Avatar Personalization}
\label{app:formalization}

This appendix provides an optional formalization of the personalization objective in Section~\ref{sec:personalization}. We use it to make precise what we mean by \emph{identity-continuous} personalization and the tension between specialization and stability.

\paragraph{Problem setting.}
We model interaction for a user $u$ over time steps $t=1,\dots,T$. At time $t$, the system observes a context $x_t$ and the user's behavior $o_t$, while the avatar produces an action $a_t \sim \pi(\cdot \mid x_t, \mathcal{I}_u^t)$. The key object of interest is the \emph{identity representation} $\mathcal{I}_u^t$, a computable proxy for the user\textquotesingle s cognition.

\paragraph{Identity representation.}
We write the avatar\textquotesingle s identity representation as
$
\mathcal{I}_u = (z_u, \mathcal{K}_u),
$
where $z_u$ captures relatively stable personality structure (e.g., values, decision style, risk preference) and $\mathcal{K}_u$ is a structured store of user-specific knowledge and experience. This decomposition separates \emph{how a user thinks} from \emph{what a user knows}.

\paragraph{Personality mining as inverse cognition modeling.}
A key challenge is to infer latent personality from observed behavior,
$z_u = f\!\left(\{(x_t, o_t)\}_{t=1}^{T}\right)$,
which can be viewed as an inverse modeling problem over cognition: rather than matching outputs locally, the goal is to recover a latent structure that renders behavior \emph{computable} and \emph{predictable} across contexts.

\paragraph{Cross-context consistency as a predictability constraint.}
A defining requirement is that the same identity should induce consistent decision structure across heterogeneous tasks. We formalize this as:

\begin{equation}
\begin{aligned}
\max_{\pi,\,\mathcal{I}_u} \quad & \mathbb{E}_{(x,o)\sim\mathcal{D}_u}\big[ \log \pi(o \mid x, \mathcal{I}_u) \big] \\
\text{s.t.} \quad & \mathcal{C}(\pi, \mathcal{I}_u) \leq \epsilon,
\end{aligned}
\end{equation}
where $\mathcal{C}(\pi,\mathcal{I}_u)$ measures identity inconsistency. Intuitively, the constraint prevents behavior from degenerating into context-local optimization.

\paragraph{Co-evolution and stability.}
Personalization is inherently dynamic: both the avatar and the user adapt over time,
$
(\mathcal{I}_u^t,\, \tilde{\mathcal{I}}_u^t) \rightarrow (\mathcal{I}_u^{t+1},\, \tilde{\mathcal{I}}_u^{t+1}).
$
To prevent drift, collapse, or unintended identity substitution, we impose a stability condition,
$
D\!\left(\mathcal{I}_u^t,\, \tilde{\mathcal{I}}_u^t\right) \le \delta,
$
which requires the learned identity to remain a faithful and stable proxy of the user.

\section{One Instantiation of AOI}
\label{app:aoi}
This appendix records one possible instantiation of the Agentic Oversight Index (AOI) used in Section~\ref{sec:aoi}. We score a concrete avatar $A$ over a task distribution $\tau$ with operational demand $\Omega$. The functional forms are illustrative; the key requirements are that net value is normalized and that human oversight is represented once, as a monotonic penalty.

\paragraph{Human oversight burden.}
Let $H(A,\tau)$ aggregate the human work needed to use, supervise, and take responsibility for an avatar:
\begin{equation}
H(A,\tau) = w_f F(A,\tau) + w_t T(A,\tau) + w_d D(A,\tau) + w_a Q(A,\tau),
\end{equation}
where $F$ is intervention frequency, $T$ is supervision time, $D$ is the cognitive complexity of required decisions, and $Q$ is accountability or audit load. The nonnegative weights encode the evaluation setting. We normalize this burden against $H_{\mathrm{human}}(\tau)$, the burden incurred when a human performs the task without an avatar:
\begin{equation}
P(A,\tau) = \frac{H(A,\tau)}{H_{\mathrm{human}}(\tau)}.
\end{equation}
Thus $P=0$ denotes no required human oversight, $P=1$ denotes human effort equal to the human-only baseline, and $P>1$ explicitly identifies an avatar that creates more human workload than a human-only workflow. Human oversight is represented exclusively by $P$, rather than also being included in agent-side operational cost.

\paragraph{Delivered value.}
Let $V(I,\Omega)$ denote the total user value produced by an agent with execution capacity $I$, decomposed into task value $A_{\mathrm{task}}(I,\Omega)$ and optional experiential value $E(I)$:
\begin{equation}
V(I,\Omega) = A_{\mathrm{task}}(I,\Omega) + E(I).
\end{equation}
We model task value using a saturation form:
\begin{equation}
A_{\mathrm{task}}(I, \Omega) = \frac{I^\alpha}{I^\alpha + \beta \cdot \Omega^\gamma} \cdot \theta(\Omega).
\end{equation}
An optional experiential term can capture satisfaction and calibrated trust without conflating them with human oversight:
\begin{equation}
E(I) = \omega \cdot \left(1 - \exp\left(-\frac{I}{I_0}\right)\right).
\end{equation}

\paragraph{Operational cost.}
We separate agent-side cost from human oversight. Operational cost comprises fixed execution consumption $C_T$ and learning cost $T_L$:
\begin{equation}
C_{\mathrm{sys}}(I,\Omega) = C_T(I,\Omega) + T_L(I),
\end{equation}
where one simple parameterization is
\begin{equation}
C_T(I, \Omega) = c_0 \cdot \Omega^{\eta} \cdot I^{-\lambda_c},
\end{equation}
\begin{equation}
T_L(I) = T_0 \cdot I^{-\rho}
\cdot \exp\left(\psi \cdot \ln\left(\frac{\bar{I}}{I}\right)\right).
\end{equation}

\paragraph{AOI.}
Given a reference scale $s_{\tau}>0$ in the same value units as $V$, $C_{\mathrm{sys}}$, and $R$, normalized net value and AOI are computed as:
\begin{equation}
U(A,\tau) = \tanh\!\left(
\frac{V(A,\tau)-C_{\mathrm{sys}}(A,\tau)}{s_{\tau}}
\right),
\end{equation}
\begin{equation}
AOI(A,\tau) = U(A,\tau) - \lambda P(A,\tau), \qquad \lambda>0.
\end{equation}

The $\tanh$ normalization keeps $U\in(-1,1)$ while preserving the ordering among different net-value outcomes. This additive form avoids the singularity of a ratio at $P=0$. Holding delivered net value fixed, it also guarantees that $\partial AOI/\partial P=-\lambda<0$, so a larger oversight burden always lowers the score, even when an agent has negative net value. Unlike a clipped $[0,1]$ participation rate, $P>1$ preserves the fact that an agent can impose more human workload than a human-only workflow.

\end{document}